\documentclass{article}

\PassOptionsToPackage{numbers,compress}{natbib}
\usepackage[final]{nips_2016}

\usepackage[utf8]{inputenc} % allow utf-8 input
\usepackage[T1]{fontenc}    % use 8-bit T1 fonts
\usepackage{hyperref}       % hyperlinks
\usepackage{url}            % simple URL typesetting
\usepackage{booktabs}       % professional-quality tables
\usepackage{amsfonts}       % blackboard math symbols
\usepackage{nicefrac}       % compact symbols for 1/2, etc.
\usepackage{microtype}      % microtypography
\usepackage{booktabs}
\usepackage{tkz-graph}
\usetikzlibrary{positioning}
\usepackage{smartdiagram}
\usesmartdiagramlibrary{additions}
\usepackage{times}
\usepackage{tabulary}
\usepackage{graphicx} % more modern
\usepackage{caption}
\usepackage{subcaption}

\usepackage{amsmath}
\usepackage{amsfonts}
\usepackage{amssymb}
\usepackage{amsthm}
\usepackage{dsfont}
\usepackage{color}
\usepackage{tikz}
\usetikzlibrary{shapes,arrows}
\usetikzlibrary{positioning}
\usetikzlibrary{positioning, fit, arrows.meta}
\usepackage{tkz-graph}
\usepackage{amsmath,bm,times}
\title{Improving LSTM-CTC based ASR  performance in domains with limited training data}

\author{
  Jayadev Billa\thanks{Author is currently unaffiliated.} \\
  \texttt{jbilla2004@gmail.com} \\
}

\begin{document}
% \nipsfinalcopy is no longer used

\maketitle

\begin{abstract}
  This paper addresses the observed performance gap between automatic speech recognition (ASR) systems based on Long Short
  Term Memory (LSTM) neural networks trained with the connectionist temporal 
  classification (CTC) loss function
  and  systems based on hybrid Deep Neural Networks (DNNs) trained with the cross entropy (CE) loss function on domains with limited data. We
  step through a number of experiments that show incremental
  improvements on a baseline EESEN toolkit based LSTM-CTC
  ASR system trained on the Librispeech 100hr ({\small \texttt{train-clean-100}}) corpus. Our 
  results show that with effective combination of data augmentation and regularization, a LSTM-CTC based system can exceed the performance of  a strong Kaldi based baseline trained on the same data. 
\end{abstract}

\setcounter{footnote}{0}
\section{Introduction}

Long Short Term Memory (LSTM),  and in general, recurrent neural network (RNN) based ASR
systems~\citep{GravesJ14,AmodeiABCCCCCCD16,HannunCCCDEPSSCN14,SakSRB15,SoltauLS16}
trained with connectionist temporal classification
(CTC)~\citep{GravesFGS06} have recently been shown to work extremely
well when there is an abundance of training data, matching and exceeding the performance of
hybrid DNN systems \citep{AmodeiABCCCCCCD16,SoltauLS16}. However, these
systems lag comparable hybrid DNN systems when trained on smaller
training sets (e.g., discussion in \citep{PoveyPGGMNWK16}). LSTM-CTC systems are much simpler to train;
they can be trained as an end-to-end speech recognition system as
opposed to the iterative multi-process approach to hybrid DNN ASR
system training. In this paper, we describe methods that eliminate the performance gap
between these two approaches and thus provide a path to wider usage of the simpler
LSTM-CTC approach in domains with lower amounts of available training data.

The work we present, in the following sections, encompasses model initialization, data augmentation, feature stacking and striding, and regularization via dropout. In particular, we show that a combination of these approaches significantly improves performance. We will first describe the key elements of our
baseline EESEN toolkit~\citep{MiaoGM15} system, then describe our efforts in the listed areas. Our results show that an LSTM-CTC based speech recognition can exceed the performance of a strong Kaldi toolkit~\citep{PoveyASRU2011} baseline trained on the same data.

\section{Baseline LSTM-CTC ASR system} \label{baselineLSTMCTC}

Our baseline system is based on the publicly available EESEN
Toolkit \citep{MiaoGM15} trained on the publicly
available Librispeech corpus \citep{PanayotovCPK15}. The acoustic model in
EESEN is a deep bidirectional LSTM neural network. In this paper, an LSTM model will always imply a bidirectional LSTM model.

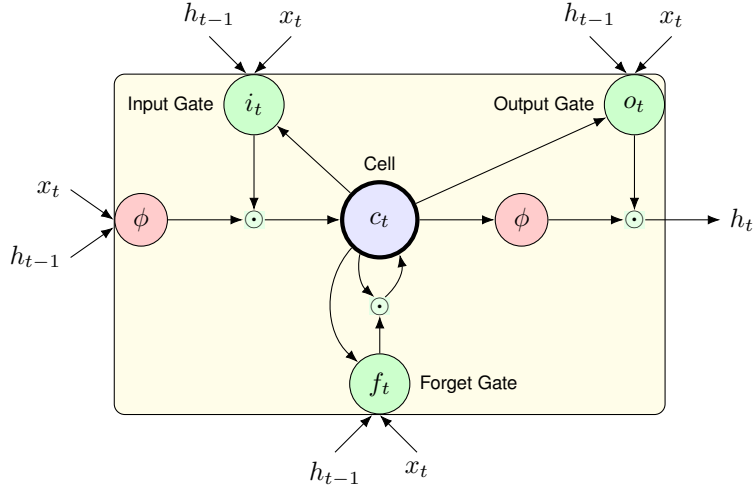
\begin{figure}[h]
  \centering
\pgfdeclarelayer{background}
\pgfdeclarelayer{foreground}
\pgfsetlayers{background,main,foreground}
\begin{tikzpicture}[
    prod/.style={fill=green!10,inner sep=0pt},
    %prod/.style={circle, draw, inner sep=0pt},
    ct/.style={circle, draw, fill=blue!10,inner sep=5pt, ultra thick, minimum width=10mm},
    ft/.style={circle, draw, fill=green!20,minimum width=8mm, inner sep=1pt},
    filter/.style={circle, draw,fill=red!20, minimum width=7mm, inner sep=1pt%, 
   % path picture={\draw[thick, rounded corners] (path picture bounding box.center)--++(65:2mm)--++(0:1mm);
    %$\draw[thick, rounded corners] (path picture bounding box.center)--++(245:2mm)--++(180:1mm);}
    },
    mylabel/.style={font=\scriptsize\sffamily},
    >=LaTeX
    ]

\node[ct, label={[mylabel]Cell}] (ct) {$c_t$};
\node[filter, right=of ct] (int1) {$\phi$};
\node[prod, right=of int1] (x1) {$\odot$}; 
\node[right=of x1] (ht) {$h_t$};
\node[prod, left=of ct] (x2) {$\odot$}; 
\node[filter, left=of x2] (int2) {$\phi$};
\node[prod, below=5mm of ct] (x3) {$\odot$}; 
\node[ft, below=5mm of x3, label={[mylabel]right:Forget Gate}] (ft) {$f_t$};
\node[ft, above=of x2, label={[mylabel]left:Input Gate}] (it) {$i_t$};
\node[ft, above=of x1, label={[mylabel]left:Output Gate}] (ot) {$o_t$};

\foreach \i/\j in {int2/x2, x2/ct, ct/int1, int1/x1,
            x1/ht, it/x2, ct/it, ct/ot, ot/x1, ft/x3}
    \draw[->] (\i)--(\j);

\draw[->] (ct) to[bend right=45] (ft);

\draw[->] (ct) to[bend right=30] (x3);
\draw[->] (x3) to[bend right=30] (ct);

    \begin{pgfonlayer}{background}
	\node[fit=(int2) (it) (ot) (ft), draw, fill=yellow!10,rounded corners, inner sep=0pt] (fit) {};
    \end{pgfonlayer}

%\draw[<-] (fit.west|-int2) coordinate (aux)--++(180:7mm) node[left]{$x_t$};
\draw[<-] (fit.west|-int2) coordinate (aux)--++(145:7mm) node[left]{$x_t$};
\draw[<-] ([yshift=-1mm]aux)--++(-145:7mm) node[left]{$h_{t-1}$};

%\draw[<-] (fit.north-|it) coordinate (aux)--++(90:7mm) node[above]{$x_t$};
\draw[<-] (fit.north-|it) coordinate (aux)--++(45:7mm) node[above]{$x_t$};
\draw[<-] ([xshift=-1mm]aux)--++(135:7mm) node[above]{$h_{t-1}$};

%\draw[<-] (fit.north-|ot) coordinate (aux)--++(90:7mm) node[above]{$x_t$};
\draw[<-] (fit.north-|ot) coordinate (aux)--++(45:7mm) node[above]{$x_t$};
\draw[<-] ([xshift=-1mm]aux)--++(135:7mm) node[above]{$h_{t-1}$};

%\draw[<-] (fit.south-|ft) coordinate (aux)--++(-90:7mm) node[below]{$x_t$};
\draw[<-] (fit.south-|ft) coordinate (aux)--++(-45:7mm) node[below]{$x_t$};
\draw[<-] ([xshift=-1mm]aux)--++(-135:7mm) node[below]{$h_{t-1}$};
\end{tikzpicture}
  \caption{LSTM Memory cell.}
  \label{lstmcell}
\end{figure}

A typical LSTM cell
is illustrated in Figure~\ref{lstmcell}. The LSTM cell consists of three gates: input, forget, and output, which
control, respectively, what fraction of the input is passed to the "memory"
cell, what fraction of the stored cell memory is retained, and what fraction of
the cell memory is output. The vector formulas that describe the LSTM
cell are
\begin{align}
    \mathbf{i}_{t} & =\sigma(\mathbf{W}_{\!i}\mathbf{x}_{t}+\mathbf{R}_{i}\mathbf{h}_{t-1}+\mathbf{P}_{i}\mathbf{c}_{t-1} + \mathbf{b}_i) & \textit{input gate}\\
    \mathbf{f}_{t} & =\sigma(\mathbf{W}_{\!\!f}\;\!\!\mathbf{x}_{t}+\mathbf{R}_{f}\:\!\!\mathbf{h}_{t-1}+\mathbf{P}_{\!f\;\!\!}\mathbf{c}_{t-1} + \mathbf{b}_{\!f\!}) & \textit{forget gate}\\
    \mathbf{c}_{t} & =  \mathbf{f}_{t}\odot\mathbf{c}_{t-1} +\mathbf{i}_{t}\odot\phi(\mathbf{W}_{\!c}\mathbf{x}_{t}+\mathbf{R}_{c}\mathbf{h}_{t-1} + \mathbf{b}_c)& \textit{cell state} \label{eqn-cell}\\
    \mathbf{o}_{t} & =\sigma(\mathbf{W}_{o}\mathbf{x}_{t}+\mathbf{R}_{o}\mathbf{h}_{t-1}+\mathbf{P}_{o}\mathbf{c}_{t}+ \mathbf{b}_o) & \textit{output gate}\\
    \mathbf{h}_{t} & =\mathbf{o}_{t}\odot \phi(\mathbf{c}_{t}) & \makebox[0pt][r]{\textit{output}}
\end{align}
where $\mathbf{x}_t$ is the input vector at time $t$, $\mathbf{W}$ are
rectangular input weight matrices connecting inputs to the LSTM cell,
$\mathbf{R}$ are square recurrent weight matrices connecting the
previous memory cell state to the LSTM cell, $\mathbf{P}$ are diagonal
peephole weight matrices and $\mathbf{b}$ are bias vectors. Functions
$\sigma$ and $\phi$ are the {\it logistic sigmoid}
($\frac{1}{1+e^{-x}}$) and {\it hyperbolic tangent} nonlinearities
respectively.  Operator $\odot$ represents the point-wise
multiplication of two vectors.

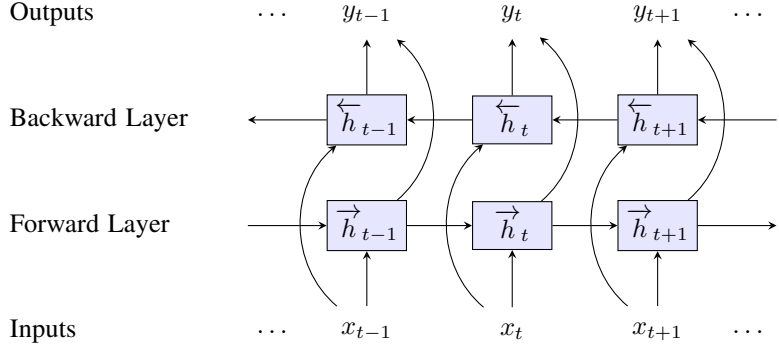
\begin{figure}
\centering

%\scriptsize
\begin{tikzpicture}[x=1em, y=1em, >=stealth,label/.style={
  rectangle,
  draw=none,
  text centered,
  inner sep=0.6em,
},
namelabel/.style={
  rectangle,
  draw=none,
  text width=12em,
  inner sep=0.6em,
  align=left
},
bigcircle/.style={
  rectangle,
  fill=blue!10,
  draw,
  minimum width=3em,
  text centered,
  inner sep=0.3em,
}]

%nodes
\node[bigcircle](bht1) at (-5.5,0){$\overleftarrow{h}_{t-1}$};
\node[bigcircle](bht2) at (0,0){$\overleftarrow{h}_{t}$};
\node[bigcircle](bht3) at (5.5,0){$\overleftarrow{h}_{t+1}$};

\node[bigcircle](fht1) at (-5.5,-4){$\overrightarrow{h}_{t-1}$};
\node[bigcircle](fht2) at (0,-4){$\overrightarrow{h}_{t}$};
\node[bigcircle](fht3) at (5.5,-4){$\overrightarrow{h}_{t+1}$};

\node[label](x0) at (-9,-8){$\dots$};
\node[label](x1) at (-5.5,-8){$x_{t-1}$};
\node[label](x2) at (0,-8){$x_{t}$};
\node[label](x3) at (5.5,-8){$x_{t+1}$};
\node[label](x4) at (9,-8){$\dots$};

\node[label](y0) at (-9,4){$\dots$};
\node[label](y1) at (-5.5,4){$y_{t-1}$};
\node[label](y2) at (0,4){$y_{t}$};
\node[label](y3) at (5.5,4){$y_{t+1}$};
\node[label](y4) at (9,4){$\dots$};

\node[namelabel](textoutputs) at (-13,4){Outputs};
\node[namelabel](textbklayer) at (-13,0){Backward Layer};
\node[namelabel](textflayer) at (-13,-4){Forward Layer};
\node[namelabel](textinputs) at (-13,-8){Inputs};

\draw[->](x1) to[bend left=50] (bht1);
\draw[->](x2) to[bend left=50] (bht2);
\draw[->](x3) to[bend left=50] (bht3);

\draw[->](fht1) to[bend right=50] (y1);
\draw[->](fht2) to[bend right=50] (y2);
\draw[->](fht3) to[bend right=50] (y3);

\draw[->] (-10,-4) -- (fht1);
\draw[->](fht1) -- (fht2);
\draw[->](fht2) -- (fht3);
\draw[->](fht3) -- (10,-4);

\draw[->] (10,0) -- (bht3);
\draw[->] (bht3) -- (bht2);
\draw[->] (bht2) -- (bht1);
\draw[->] (bht1) -- (-10,0);

\draw[->] (x1) -- (fht1);
\draw[->] (x2) -- (fht2);
\draw[->] (x3) -- (fht3);

\draw[->] (bht1) -- (y1);
\draw[->] (bht2) -- (y2);
\draw[->] (bht3) -- (y3);

\end{tikzpicture} 
\caption{Bidirectional LSTM Layer, boxes represent a single LSTM memory cell.}
\label{blstm-layer}
\end{figure}

These cells are then arranged into a bidirectional layer where data is
processed independently in the forward and backward directions~\cite{SchusterP97}. The
outputs from both forward and backward directions are concatenated to
form the input to the next recurrent layer as illustrated in
Figure~\ref{blstm-layer}.
\begin{align}
	\mathbf{y}_t = [\overrightarrow{\mathbf{h}}_{t}, \overleftarrow{\mathbf{h}}_{t}]
\end{align}

The final model is then composed of multiple stacked bidirectional
LSTM layers, resulting in a deep bidirectional LSTM neural network
followed by a softmax layer with $K+1$ outputs nodes, where $K$ is the count
of output labels with an additional node to represent a \emph{blank}
label $\phi$; the blank label allows the network to output a "none"
 label when evidence is insufficient to output a specific label. The bidirectional LSTM network is trained with the CTC loss
function, which minimizes the negative log summed probability of the
correct label sequence given the input, via a forward-backward
algorithm that sums across all possible alignments. Details can be
found in \citep{GravesFGS06}. Unless otherwise indicated, the LSTM
  model in our experiments consists of 4 bidirectional stacked layers
  of 640 LSTM cells (320 in each direction).

Our system training procedure mirrors the EESEN WSJ recipe;
inputs are 40 dimensional mel warped filterbank features with $\Delta$
and $\Delta\Delta$ features, normalized with per speaker means
subtraction and variance normalization, and outputs are the grapheme
or phoneme sequence. EESEN uses a weighted finite state grammar (WFST) to incorporate the language model and generate the final word sequence from the network outputs. Details can be found in \cite{MiaoGM15}.

During training, we use the ``newbob'' learning rate (LR) schedule,
where the LR is first set to a fixed value (0.00004) and halved every
epoch once the improvement in token accuracy on a withheld
cross-validation (CV) set falls below 0.5\%. Training is stopped once
the improvement in CV token accuracy falls below
0.1\%. In later experiments, discussed in this paper, we modified this approach to allow training to continue until the token accuracy had either plateaued over several epochs or was clearly not going to exceed that of other methods, before triggering LR halving.

We choose the Librispeech 100hr ({\small \texttt{train-clean-100}}) corpus as our training set. This corpus is publicly available and allows others to easily reproduce our results using our modified EESEN code available at  {\small \texttt{https://github.com/jb1999/eesen}}. Table~\ref{baseline} presents the results of this system on the Librispeech 
\emph{dev-clean} test set when trained on grapheme and phoneme output units. Two
variants of the phoneme system were trained, one that used the phoneme
set provided with the Librispeech corpus that includes phoneme stress
markers, and another where the stress markers were removed creating a reduced phoneme set. For the phoneme based systems, we use the Librispeech provided dictionary, but remove any alternate pronunciations so that each word has only one pronunciation. 

In all experiments, the
Librispeech supplied 3-gram ARPA LM, pruned with threshold 1e-7, {\small \texttt{tgmed}} in the recipe, is
used as the language model during decoding. Given that the reduced phoneme model showed the best performance across the trained EESEN systems, we adopted this variant as our canonical model, all subsequent experiments in this paper incrementally add capabilities to this system.

\begin{table}[t]
  \caption{Baseline results using Librispeech 100hr  ({\small \texttt{train-clean-100}}) corpus as training data and pruned 3-gram ARPA LM ({\small \texttt{tgmed}}). For comparison, the last two rows show performance when Librispeech 360hr ({\small \texttt{train-clean-360}}) and Librispeech 500hr ({\small \texttt{train-clean-500}}) training sets are incrementally added.}
  \label{baseline}
  \centering
  \begin{tabular}{lc}
    \toprule
    System     &  \%WER (\emph{dev-clean}) \\
    \midrule
    Grapheme                         &  13.01 \\ %e17
    Phoneme                          &  11.81 \\ %e16
    Reduced Phoneme                 &  11.15 \\ %e19
    Kaldi (p-norm DNN, LDA+MLLT+SAT, 100hrs training)\textsuperscript{\dag} &   7.91 \\
    \midrule
    Kaldi (p-norm DNN, LDA+MLLT+SAT, 460hrs training)\textsuperscript{\dag} &   7.16 \\
    Kaldi (p-norm DNN, LDA+MLLT+SAT, 960hrs training)\textsuperscript{\dag} &   6.57 \\    
    \bottomrule
    \multicolumn{2}{l}{\textsuperscript{\dag} \scriptsize{\texttt{https://github.com/kaldi-asr/kaldi/blob/master/egs/librispeech/s5/RESULTS} (retrieved 5/30/2017).}}
  \end{tabular}
\end{table}

Also shown in Table~\ref{baseline} is the
Kaldi baseline performance following the s5 recipe for Librispeech
100hrs with the same language model. This Kaldi model uses composite mel frequency cepstral coefficient (MFCC) features over which Linear Discriminant Analysis (LDA),  Maximum Likelihood Linear Transform (MLLT) and Speaker Adaption Transform (SAT) transformations are applied to generate 40 dimensional features used during training; the DNN architecture itself consists of 4 p-norm layers with 3486 outputs corresponding to the context dependent clusters.
We should note here that the Kaldi baseline system in Table~\ref{baseline} has about half the trainable parameters, 4.5M parameters, as opposed to the corresponding EESEN systems with 9.1M parameters. We did increase the size of the Kaldi baseline model by proportionally adjusting the p-norm layer input and output dimensions to approximately match the EESEN system trainable parameters. However, the larger Kaldi system's performance was worse, with a 9.02\% WER, albeit with no specific tuning/optimization on our part.  Note also that the Kaldi baseline incorporates context dependency modeling whereas the EESEN models output context independent phonemes/graphemes.  

\section{LSTM Initialization}

Typically, LSTM parameters are initialized to small random values in an
interval, e.g., in our experiments we initialize weights and biases to random
values in the interval $[-0.1,0.1]$. However, for the forget gate, this is a suboptimal
choice, where small weights effectively close the gate, preventing cell memory and its gradients from flowing in time. 
One approach to address this issue is to initialize the
forget gate bias, $\mathbf{b}_{f}$, to a large value, say $1$, that
will force the gate to be initialized in an open position and allow
memory cell gradients in time to flow more readily. This practical
detail has been noted in earlier
work \citep{GersSC00,JozefowiczZS15,MiaoGNKMW16}. In our work,
however, we found that simply increasing the forget gate bias did not
provide the anticipated gains, and during training, accuracy in early
epochs plateaued very quickly and triggered our stopping criterion,
with worse results than if we had left the forget gate bias set to a
random small initial value. However, if we disable the stopping criterion and force a minimum number of
epochs, in this case 6 epochs, we did
observe consistent gains with $\mathbf{b}_{f}=\mathbf{1}$. The results from our experiments with forget gate bias initialization are summarized in Table~\ref{fgresults}.

\begin{table}[t]
  \caption{Forget gate bias initialization.}
  \label{fgresults}
  \centering
  \begin{tabular}{lc}
    \toprule
    System     &  \%WER (\emph{dev-clean}) \\
    \midrule
    Random initialization &  11.15 \\ %e19
    $\mathbf{b}_{f}=\mathbf{1}$  &  12.23 \\ %e43
    $\mathbf{b}_{f}=\mathbf{1}$ $+$ minimum 6 epochs before LR halving  &  10.76 \\ %e44
    \bottomrule
  \end{tabular}
\end{table}

As a practical matter, there is no reliable way to predetermine the minimum epochs for any particular system. In later experiments, we opted to simply disable the stopping criterion, monitor accuracy on our withheld cross validation set, and manually trigger ``newbob'' learning rate (LR) schedule, when the token accuracy had plateaued over several epochs.

\section{Data Augmentation} 

Data augmentation is used in state of the art ASR systems to increase the available training data, improve generalization capability, as well as improve system robustness to the deployment environment. Typically, data is augmented by either adding noise from other sources \citep{HannunCCCDEPSSCN14,SakSRB15} or by explicit transformation of the data either in the time domain as speed perturbation \citep{KoPPK15} or frequency domain warping using vocal tract length perturbation (VTLP) \citep{jaitly2013,CuiGK15}. In the first case, we need a suitable noise source and method to add the appropriate noise to the training data. For speed perturbation, we need an external process that modifies the data prior to feature extraction. In VTLP, either the utterance is warped with a random factor \citep{jaitly2013},  or the existing warp factor for the speaker is jittered \citep{CuiGK15}. Another variation, Stochastic Feature Mapping~\citep{CuiGK15}, seeks to augment data by converting one speaker's speech to another speaker. 

An alternate perspective is to view data augmentation as an exploration method to push the model to explore and find a more beneficial space on the manifold, which in turn, provides a path to a deeper minimum. This is conceptually similar to simulated annealing \citep{Kirkpatrick1983}, in that we consider data augmentation as analogous to changing temperatures to allow for more robust exploration for a global minimum. Since we are looking to encourage model exploration vs generalization, our implementation of data augmentation diverges from a more typical implementation in two ways:
\begin{itemize}
\item Data augmentation is applied to introduce data variety without attempting to faithfully retain audio characteristics. In our case, we choose to uniformly warp all the data in the frequency domain using VTLN warp factors 0.8 and 1.2 across all the training data, in addition to retaining the original unwarped features. This clearly is wrong in many cases, where these warp factors result in either excessive compression or expansion of the frequency spectrum. To provide an additional level of variance,  we also modified the frame rate to speed up or slow down the features. We use frame rates of 8ms and 11ms in addition to the original 10ms frame rate.
\item We train the model with different set of augmented data every epoch and cycle through the different augmented data sets, as opposed to selecting from a different augmented data set for every mini-batch or utterance.
\end{itemize}

We refer to this collective approach to data augmentation as \emph{max} perturbation to distinguish from approaches that attempt to more faithfully retain the audio characteristics and seek to introduce small perturbations during training. Further, this approach has the advantage of leveraging existing feature extraction tools and training process, albeit with different parameter settings, without requiring additional tools or processes. 

Table~\ref{dataaugres} summarizes the gains from using max perturbation in comparison with the speed perturbation approach, which in earlier work has been shown to provide better performance than frequency based transformation~\citep{KoPPK15}. With 9-fold max perturbation, we observed a 9.1\% relative reduction in WER, whereas 3-fold speed perturbation as described in \citep{KoPPK15} but with per epoch augmented training described above, resulted in a 5.0\% relative reduction in WER. For comparison, we note that Ko et. al. in \citep{KoPPK15} show speech perturbation results across a variety of data sets with relative reduction in WER ranging from 6.7\% to 0.32\% .  Our speed perturbation result of 5.0\% relative improvement is consistent with, and in the high end of, these prior results.  In other experiments (not shown here), we observed that ~75\% of the max perturbation gain can be attributed to VTLN warping data augmentation and ~25\% of the gain to frame rate data augmentation, and that both these gains were orthogonal and largely additive.

In order to explore how max perturbation behaves with more data variants, we also experimented with 20-fold max perturbation with frame rates 8, 10,11 and 12ms, and VTLN warp factors 0.7, 0.8, 1.0, 1.2 and 1.3. We found that 20-fold max perturbation showed an improvement over no data augmentation and 3-fold speed perturbation, but was slightly worse than 9-fold max perturbation with a 7.5\% relative reduction in WER over no data augmentation. This result is quite interesting considering that we are well outside of traditional norms in terms of how much we can distort speech data and still retain modeling value. 

\begin{table}[t]
  \caption{Improvements from data augmentation.}
  \label{dataaugres}
  \centering
  \begin{tabular}{lc}
    \toprule
    System     &  \%WER (\emph{dev-clean}) \\
    \midrule
    prior result with $\mathbf{b}_{f}=\mathbf{1}$  &  10.76 \\ %e44
    9-fold max perturbation  &  9.78  \\  %e72 (10.76-9.78)/10.76 =  0.0911
    20-fold max perturbation & 9.95 \\  %e136 (10.76-9.95)/10.76 = 0.0753
    3-fold speed perturbation & 10.22 \\ %e135 (10.76-10.22)/10.76 = 0.0502
    \bottomrule
  \end{tabular}
\end{table}  
 
Given the relative outperformance of max perturbation data augmentation vis-a-vis speed perturbation we used 9-fold max perturbation in all of our subsequent experiments.

\section{Stacking  and striding frames} \label{ssf}
Frame stacking is where consecutive feature frames are concatenated to create a composite feature -- this increases the size of the input feature. Frame striding is where feature frames at particular times are skipped and not presented to the network -- this is equivalent to downsampling the feature frame sequence. Both approaches have been used independently and concurrently in prior DNN and LSTM-CTC systems \citep{AmodeiABCCCCCCD16,SakSRB15,PoveyASRU2011} to better capture discriminative features as well as to reduce computation and system training time. 

We conducted several experiments with stacking and striding, and observed a consistent drop in performance. Illustrative results are shown in Table~\ref{framestack}, with  $\pm1$ feature frames, for a composite stacked feature of 3 frames, and a stride of 3, for an effective 30ms frame rate vs. the original 10ms frame rate. We find that 9-fold max perturbation data augmentation provides a slight, but relatively stronger, performance improvement (10.1\%  vs. 9.1\% relative) when compared to the model based on the base features. 
\begin{table}[t]
  \caption{Experiments with frame stacking and striding.}
  \label{framestack}
  \centering
  \begin{tabular}{lc}
    \toprule
    System     &  \%WER (\emph{dev-clean}) \\
    \midrule
    Baseline &  10.76 \\ %e44
    Baseline + 9-fold max perturbation &  9.78 \\  %e72 (10.76-9.78)/10.76 =  0.0911
    3 frame stack, stride $=3$  &  11.92  \\  %e74 
    3 frame stack, stride $=3$ + 9-fold max perturbation  &  10.72  \\  %e75 (11.92-10.72)/11.92=0.1007
    \bottomrule
  \end{tabular}
\end{table} 
While these results are not compelling by themselves, as we show in the next section, stacking and striding is a key element in driving performance improvements with dropout in LSTM-CTC systems.

\section{Dropout}
In many of our initial experiments, we observed consistent overfitting, as also noted by others \citep{SakSB14}. In order to minimize overfitting we explored various approaches to dropout on feedforward and recurrent connections. While the general approach to apply dropout is well established for feedforward networks \cite{SrivastavaHKSS14}, application to recurrent neural networks, however, has seen a number of variants~\citep{ZarembaSV14,MoonCLS15,SemeniutaSB16,GalG16}. 

Initial implementations of dropout for recurrent neural networks focused on application of dropout on the feedforward connections only~\citep{ZarembaSV14, PhamBKL14}. In subsequent work, Moon et. al.~\citep{MoonCLS15} introduced RNNDrop, in which dropout is applied to LSTM cell memory at a sequence level, i.e., the dropout mask is kept fixed across each training sequence, or in the case of speech, each utterance. Gal et. al.~\citep{GalG16} Variational RNN approach is similar, but dropout is applied to the LSTM cell output recurrent connection as well as the forward non-recurrent connections, also at a sequence level. An interesting variant is  recurrent dropout without memory loss~\citep{SemeniutaSB16}, where dropout is applied to the LSTM cell memory update, which prevents the LSTM cell memory from being reset as is the case with RNNDrop. More recently, Cheng et. al.~\citep{ChengPPMKY17} explore dropout in projected LSTM based ASR systems. Given the slightly different architecture of the projected LSTM cell, they investigate different dropout location approaches and further incorporate a schedule driven dropout rate, where the dropout factor changes over time and show relative WER reductions on the order of 3\% to 7\% on a variety of data sets.

In our experiments, detailed below, we systematically explore the application of dropout to the feedforward and recurrent connections, separately, as well as together. We show that careful application of dropout, when coupled with max perturbation and stacked and strided features, is a significant driver of system performance. 

We should note that in our implementation of dropout, we scale the mask during training with no change to the test paradigm. To maintain a common operating point for our experiments, we fix the dropout rate to $0.2$ for forward and recurrent connections as applicable.  It is likely that a more comprehensive exploration of dropout rates will provide additional performance improvements.

\subsection{Dropout on feedforward connections}

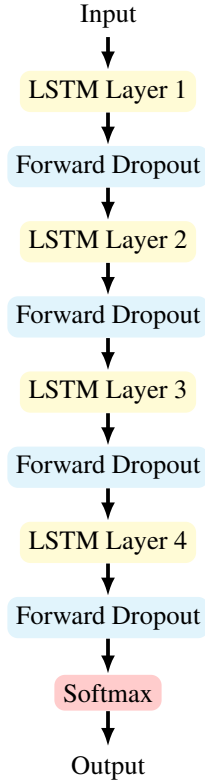
\begin{figure}
\centering
\usetikzlibrary{shapes,arrows}
\usetikzlibrary{positioning}
\usetikzlibrary{positioning, fit, arrows.meta}
\tikzset{%
 % >={Latex[width=2mm,length=2mm]},
  % Specifications for style of nodes:
            base/.style = {rectangle, rounded corners, %draw=black,
                           text centered},
             end/.style = {%rectangle, rounded corners, draw=black,
                           text centered},                          
    lstm/.style = {base, fill=yellow!20},
    dropout/.style = {base, fill=cyan!10},
    softmax/.style = {base, fill=red!20},       
    startstop/.style = {end},
}
\begin{tikzpicture}[%node distance=1.5cm,
    every node/.style={fill=white}, align=center]
  % Specification of nodes (position, etc.)
  \node (in)             [end]              {Input};
  \node (layer1)   [lstm, below of=in]             {LSTM Layer 1};
  \node (dlayer1) [dropout, below of=layer1] {Forward Dropout};  
  \node (layer2)   [lstm, below of=dlayer1]    {LSTM Layer 2};
  \node (dlayer2) [dropout, below of=layer2] {Forward Dropout};
  \node (layer3)   [lstm, below of=dlayer2]    {LSTM Layer 3};
  \node (dlayer3) [dropout, below of=layer3] {Forward Dropout};
  \node (layer4)   [lstm, below of=dlayer3]     {LSTM Layer 4};
  \node (dlayer4) [dropout, below of=layer4]  {Forward Dropout};
  \node (layer5)   [softmax, below of=dlayer4] {Softmax};
  \node (out) [end, below of=layer5]      {Output};     
  % Specification of lines between nodes specified above
  % with aditional nodes for description 
  \draw[line width=0.5mm, -{Latex[length=2.5mm]}]       (in) -- (layer1);
  \draw[line width=0.5mm, -{Latex[length=2.5mm]}]     (layer1) -- (dlayer1);
  \draw[line width=0.5mm, -{Latex[length=2.5mm]}]     (dlayer1) -- (layer2);
  \draw[line width=0.5mm, -{Latex[length=2.5mm]}]     (layer2) -- (dlayer2);
  \draw[line width=0.5mm, -{Latex[length=2.5mm]}]     (dlayer2) -- (layer3);
  \draw[line width=0.5mm, -{Latex[length=2.5mm]}]     (layer3) -- (dlayer3);
  \draw[line width=0.5mm, -{Latex[length=2.5mm]}]     (dlayer3) -- (layer4);
  \draw[line width=0.5mm, -{Latex[length=2.5mm]}]     (layer4) -- (dlayer4);
  \draw[line width=0.5mm, -{Latex[length=2.5mm]}]     (dlayer4) -- (layer5);   
  \draw[line width=0.5mm, -{Latex[length=2.5mm]}]     (layer5) -- (out);
\end{tikzpicture}
\caption{Forward Dropout as implemented in our experiments. The dropout mask can be sampled either every time step or every sequence. In the latter,  the dropout mask is fixed for all time steps in any specific sequence.}
\label{fdrop}
\end{figure}

Dropout on feedforward connections in LSTM networks is closely aligned with the original formulation of dropout where the composite LSTM cell is the unit to be dropped. In earlier work, Pham et. al. \cite{PhamBKL14} and Zaremba et. al.  \citep{ZarembaSV14}, make the argument for applying dropout only on the feedforward connections and not the recurrent connections so as to minimize impact on the sequence modeling capability of a recurrent network. In the implementation described in  \citep{ZarembaSV14}, dropout is applied every time step to the feedforward connections. However, given the within layer recurrence, application of dropout at every time step would be a noisy implementation of the classical dropout approach, i.e., each sampling of the networks would receive input from another sampling of the network via the recurrent connections. We argue that a more faithful implementation for recurrent networks would be to retain the dropout mask across a complete sequence/utterance for the forward non-recurrent connections to eliminate this cross-sampling noise. In this instance, each sequence is trained on a particular sampling of the network, rather than a multitude of sampled networks driving a noisy recurrence.  We note that this is identical to the Gal et. al. \citep{GalG16} dropout implementation for forward connections. Figure~\ref{fdrop} illustrates the forward dropout implementation in our experiments.

In order to validate our thinking, we experimented with both time step and sequence forward dropout variants. Table~\ref{forwarddp} details our results. We expected a performance improvement with dropout mask sampled every time step and a larger improvement when the dropout mask is sampled every sequence. However, while we see an improvement with dropout mask sampling every time step, when the dropout mask is sampled every sequence, performance is worse than without dropout. Inspection of the training logs showed that this model started to overfit early, driving the lower performance.
\begin{table}[t]
  \caption{Experiments with dropout on feedforward connections. Forward-step indicates feedforward connection dropout with dropout mask sampled every time step. Forward-sequence indicates feedforward connection dropout with dropout mask sampled every sequence.}
  \label{forwarddp}
  \centering
  \begin{tabular}{lc}
    \toprule
    System     &  \%WER (\emph{dev-clean}) \\
    \midrule
    Baseline (9-fold max perturbation) &  9.78 \\ %e72
    Forward-step & 9.51 \\ %e138
    Forward-sequence & 10.03 \\ %e139
    \bottomrule
  \end{tabular}
\end{table} 

Given the slightly better relative WER improvement with 9-fold max perturbation using stacked and strided features, we proceeded to investigate if dropout would also show a similar magnified effect when coupled with stacked and strided features. We applied the same forward dropout variants as in Table~\ref{forwarddp} with stacked and strided features; Table~\ref{forwarddpss} details these results.  Once again, we see a larger improvement when using stacked and strided features; models trained with stacked and strided features are much more responsive to dropout, showing a significant improvement in performance over models trained with the base features. Per time step dropout mask sampling with 9-max perturbation yields a 14.5\% relative WER improvement over the corresponding baseline, and per sequence dropout mask sampling yields 19.5\% relative WER improvement. To be conservative, compared to the base feature system with 9-fold max perturbation with 9.78\% WER, we still see a relative WER improvement of 6.2\% and 11.8\% respectively for the time step and sequence variants.
\begin{table}[t]
  \caption{Experiments with dropout on feedforward connections with stacked/strided features.}
  \label{forwarddpss}
  \centering
  \begin{tabular}{lc}
    \toprule
    System     &  \%WER (\emph{dev-clean}) \\
    \midrule
    Baseline (9-fold max perturbation + stacked/strided features)  &  10.72 \\ %e75
    Forward-step & 9.17 \\ %e140 (10.72-9.17)/10.72 = 0.1446 ; (9.78-9.17)/9.78 = 0.0624
    Forward-sequence & 8.63 \\ %e141 (10.72-8.63)/10.72 = 0.1950; (9.78-8.63)/9.78 = 0.1176
    \bottomrule
  \end{tabular}
\end{table}

During training we found that networks trained on stacked and strided features were able to train for many more epochs without overfitting when coupled with dropout, resulting in performance that is significantly better than using the base features with/without dropout or the stacked and strided features without dropout. Given the strong outperformance of stacked and strided features with dropout we shifted to using stacked and strided features for all of our subsequent dropout experiments. 

\subsection{Dropout on Recurrent connections}
In our exploration of dropout for recurrent connections, we looked at two versions of recurrent connection dropout, RNNDrop~\citep{MoonCLS15} and recurrent dropout without memory loss~\citep{SemeniutaSB16}. As described earlier, RNNDrop applies dropout to the memory cell content, in particular, Equation~\ref{eqn-cell} which describes the memory cell, changes as below
\begin{align}
    \mathbf{c}_{t} & =   \mathfrak{m}_{t} \odot \big(\mathbf{f}_{t}\odot\mathbf{c}_{t-1} +\mathbf{i}_{t}\odot\phi(\mathbf{W}_{\!c}\mathbf{x}_{t}+\mathbf{R}_{c}\mathbf{h}_{t-1} + \mathbf{b}_c)\big) 
\end{align}
where $\mathfrak{m}_{t}$ represents the dropout mask at time $t$.

In the case of recurrent dropout without memory loss, dropout is only applied to the incremental memory cell update, Equation~\ref{eqn-cell} changes as below
\begin{align}
    \mathbf{c}_{t} & =   \mathbf{f}_{t}\odot\mathbf{c}_{t-1} +\mathfrak{m}_{t} \odot \mathbf{i}_{t}\odot\phi(\mathbf{W}_{\!c}\mathbf{x}_{t}+\mathbf{R}_{c}\mathbf{h}_{t-1} + \mathbf{b}_c) 
\end{align}
where again $\mathfrak{m}_{t}$ represents the dropout mask at time $t$.

We expect recurrent dropout without memory loss to show better performance vis-\`{a}-vis RNNDrop since the cell memory is not being reset continually as is the case with RNNDrop.

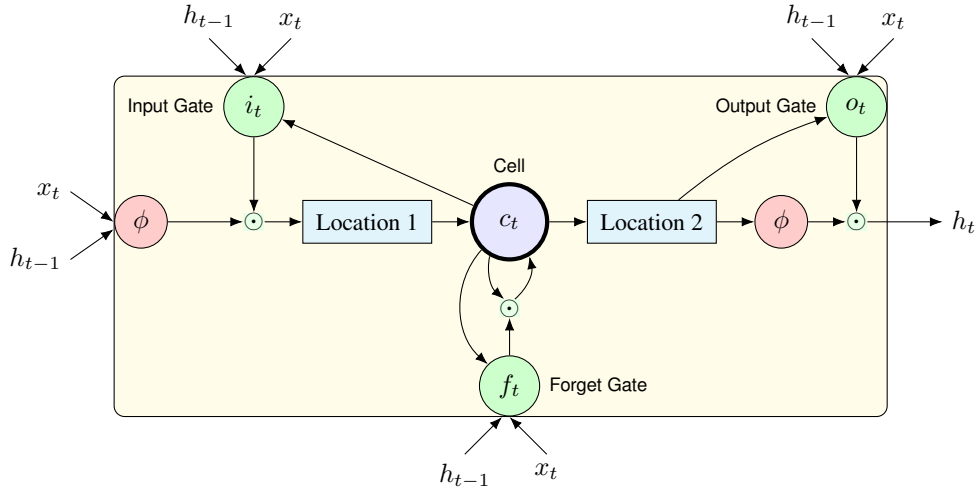
\begin{figure}[h]
  \centering
\pgfdeclarelayer{background}
\pgfdeclarelayer{foreground}
\pgfsetlayers{background,main,foreground}
\begin{tikzpicture}[
    prod/.style={fill=green!10,inner sep=0pt},
    %prod/.style={circle, draw, inner sep=0pt},
    dp/.style={rectangle, draw, fill=cyan!10,inner sep=5pt, minimum width=10mm},
    ct/.style={circle, draw, fill=blue!10,inner sep=5pt, ultra thick, minimum width=10mm},
    ft/.style={circle, draw, fill=green!20,minimum width=8mm, inner sep=1pt},
    filter/.style={circle, draw,fill=red!20, minimum width=7mm, inner sep=1pt%, 
   % path picture={\draw[thick, rounded corners] (path picture bounding box.center)--++(65:2mm)--++(0:1mm);
    %$\draw[thick, rounded corners] (path picture bounding box.center)--++(245:2mm)--++(180:1mm);}
    },
    mylabel/.style={font=\scriptsize\sffamily},
    >=LaTeX
    ]

\node[ct, label={[mylabel]Cell}] (ct) {$c_t$};
\node[dp, right=5mm of ct] (dp1) {{\footnotesize Location 2}};
\node[dp, left=5mm of ct] (dp2) {{\footnotesize Location 1}};
\node[filter, right=5mm of dp1] (int1) {$\phi$};
\node[prod, right=5mm of int1] (x1) {$\odot$}; 
\node[right=of x1] (ht) {$h_t$};
\node[prod, left=5mm of dp2] (x2) {$\odot$}; 
\node[filter, left=of x2] (int2) {$\phi$};
\node[prod, below=5mm of ct] (x3) {$\odot$}; 
%\node[dp, right=4mm of x3] (dp2) {{\footnotesize Location 2}};
\node[ft, below=5mm of x3, label={[mylabel]right:Forget Gate}] (ft) {$f_t$};
\node[ft, above=of x2, label={[mylabel]left:Input Gate}] (it) {$i_t$};
\node[ft, above=of x1, label={[mylabel]left:Output Gate}] (ot) {$o_t$};

\foreach \i/\j in {int2/x2, x2/dp2, ct/dp1,dp1/int1, int1/x1, dp2/ct,
            x1/ht, it/x2, ct/it,  ot/x1, ft/x3}
    \draw[->] (\i)--(\j);

\draw[->] (ct) to[bend right=45] (ft);
\draw[->] (dp1) to[bend left=10] (ot);
\draw[->] (ct) to[bend right=30] (x3);
\draw[->] (x3) to[bend right=30] (ct);
%\draw[->] (dp2) to[bend right=20] (ct);

    \begin{pgfonlayer}{background}
	\node[fit=(int2) (it) (ot) (ft), draw, fill=yellow!10,rounded corners, inner sep=0pt] (fit) {};
    \end{pgfonlayer}

%\draw[<-] (fit.west|-int2) coordinate (aux)--++(180:7mm) node[left]{$x_t$};
\draw[<-] (fit.west|-int2) coordinate (aux)--++(145:7mm) node[left]{$x_t$};
\draw[<-] ([yshift=-1mm]aux)--++(-145:7mm) node[left]{$h_{t-1}$};

%\draw[<-] (fit.north-|it) coordinate (aux)--++(90:7mm) node[above]{$x_t$};
\draw[<-] (fit.north-|it) coordinate (aux)--++(45:7mm) node[above]{$x_t$};
\draw[<-] ([xshift=-1mm]aux)--++(135:7mm) node[above]{$h_{t-1}$};

%\draw[<-] (fit.north-|ot) coordinate (aux)--++(90:7mm) node[above]{$x_t$};
\draw[<-] (fit.north-|ot) coordinate (aux)--++(45:7mm) node[above]{$x_t$};
\draw[<-] ([xshift=-1mm]aux)--++(135:7mm) node[above]{$h_{t-1}$};

%\draw[<-] (fit.south-|ft) coordinate (aux)--++(-90:7mm) node[below]{$x_t$};
\draw[<-] (fit.south-|ft) coordinate (aux)--++(-45:7mm) node[below]{$x_t$};
\draw[<-] ([xshift=-1mm]aux)--++(-135:7mm) node[below]{$h_{t-1}$};
\end{tikzpicture}
  \caption{LSTM Memory cell with recurrent dropout locations.  Location 1 corresponds to dropout location for recurrent dropout without memory loss, Location 2 corresponds to dropout location for RNNDrop.}
  \label{lstmcellrdp}
\end{figure}

Figure~\ref{lstmcellrdp} illustrates the location of these two recurrent dropout approaches. Location 1 corresponds to the application of recurrent dropout without memory loss, Location 2 corresponds the application of RNNDrop.

As we did with forward dropout, we decided to explore both per time step and per sequence implementation variants on both RNNDrop and recurrent dropout without memory loss. However, during our experimentation, we discovered that the direct implementation of RNNDrop as described in \citep{MoonCLS15} with per sequence dropout mask did not train and suffered from an exploding memory cell value problem, an issue predicted in \citep{SemeniutaSB16}, and was subsequently excluded from our experiments.

Table~\ref{recurrentdp} details our results, where we have abbreviated recurrent dropout without memory loss as no memory loss (NML) dropout. From the table it is clear that application of dropout for recurrent connections is indeed beneficial. In line with our expectations, NML dropout worked better than RNNDrop with the per sequence dropout mask variant slightly edging out the per time step dropout mask variant, inline with trends reported in \citep{SemeniutaSB16} albeit on different tasks.  For the NML dropout variants we see relative WER reductions over 20\% compared to the corresponding baseline, or $\sim$13\% relative WER reduction compared to the prior best system without dropout WER of 9.78\%.
\begin{table}[t]
  \caption{Experiments with dropout on recurrent connections. Postfix step and sequence indicates whether dropout mask is sampled every time step or every sequence. Recurrent dropout without memory loss is abbreviated as no memory loss (NML) dropout.}
  \label{recurrentdp}
  \centering
  \begin{tabular}{lc}
    \toprule
    System     &  \%WER (\emph{dev-clean}) \\
    \midrule
    Baseline (9-fold max perturbation + stacked/strided features) &  10.72 \\ %e75
    RNNDrop-step & 9.07 \\ %e115
    NML-step & 8.55 \\ %e116 (10.72-8.55)/10.72 = 0.2024; (9.78-8.55)/9.78 = 0.1258
    NML-sequence & 8.45 \\ %e119
    \bottomrule
  \end{tabular}
\end{table} 
 
We note that in the case of RNNDrop implemented with a per time step mask, with a dropout rate of $0.2$,  we effectively reset ${\sim}$99\%\footnote{ $1-0.8^{20}$} of all LSTM cells within 20 time steps, i.e., each LSTM cell retains at most 600ms of memory with our 30ms frame rate. An interesting corollary of this observation is that we can consider refactoring a bidirectional LSTM with RNNDrop system as an unidirectional LSTM system with suitably delayed output, with similar or near similar performance and lower latency.

\subsection{Combining feedforward and recurrent dropout}

In the previous sections we have shown that both forward and recurrent dropout mechanisms improve system performance to a similar degree. The next logical extension is to combine both forward and recurrent dropout in a single model and explore if these performance gains are complementary. 

\subsubsection{Na\"{\i}ve dropout combination}

The simplest approach to combine dropout is to apply both forward and recurrent dropout concurrently during network training; we refer to this combined dropout approach as \textit{na\"{\i}ve} dropout combination. Indeed this is the approach that has been traditionally taken while combining dropout~\citep{GalG16,SemeniutaSB16}. 

To be exhaustive we ran experiments with all combinations of the three recurrent dropout variants and the two forward dropout variants, albeit with the same fixed dropout rate of $0.2$.  Table~\ref{naivecomb} summarizes these models and their corresponding WERs. The RNNDrop combination variants, while showing better WER performance than the RNNDrop alone system, are worse, or at best similar, in performance to the forward dropout alone models. On the other hand, the NML combination variants all show performance improvements over the NML or forward dropout alone models. We note that with these results, the LSTM-CTC system is at parity with the Kaldi Librispeech 100hr hybrid DNN system in Table~\ref{baseline}.

\begin{table}[t]
  \caption{Experiments with  Na\"{\i}ve dropout combination.}
  \label{naivecomb}
  \centering
  \begin{tabular}{lccc}
    \toprule
    System &  \%WER (\emph{dev-clean})\\
    \midrule
    RNNDrop-step + Forward-step & 8.60 \\ %e142
    RNNDrop-step + Forward-sequence &  8.85\\ %e143, %e122
    NML-step + Forward-step &  8.08\\ %e147, e126
    NML-step + Forward-sequence &  7.76\\ %e145, e124
    NML-sequence + Forward-step & 7.72\\ %e146, e125
    NML-sequence + Forward-sequence& 7.97\\ %e144, e123
    \bottomrule
  \end{tabular}
\end{table}

\subsubsection{Stochastic and cascade dropout combination}

During our experimentation, it was clear that different dropout approaches had very different training profiles in how token accuracy improved on training and cross-validation and the corresponding difference in metrics. One conjecture is that the different dropout approaches explore a different part of the parameter space. If this is the case, we can direct this exploration more intelligently than a na\"{\i}ve dropout combination. 

One combination approach is to apply forward or recurrent dropout singly rather than concurrently. The exact approach we implemented is that for each minibatch, we pick from an equiprobable Bernoulli distribution to decide between forward or recurrent dropout, and then apply the appropriate dropout for that minibatch. Note that there is nothing special in our choice of distribution or decision choice, an equally valid implementation could bias towards a particular dropout or introduce an additional decision choice to pick between forward and/or recurrent dropout types. We refer to this general approach of distribution based choice to determine dropout combination as \textit{stochastic} dropout combination.

Another combination approach is to train the model with one type, combination or parameterization of dropout, and then switch to a different type of dropout, combination or parameterization, at an opportune time. Given that we cascade different dropout combinations during training we refer to this general approach as \textit{cascade} dropout combination. The dropout schedule approach described in Cheng et. al.~\citep{ChengPPMKY17} is a specific example of cascade dropout combination. We should note that cascade and stochastic dropout combination are orthogonal approaches and can be applied at the same time.

Given our limited compute resources, we were unable to systematically explore the many permutations of dropout combination in a reasonable time frame. Table~\ref{stocomb} summarizes our experiments with stochastic dropout combination and compares the results with na\"{\i}ve dropout combination. We find that while not all stochastic dropout combination results show better performance over na\"{\i}ve dropout combination, in the case of sequence based NML/Forward dropout combination we see an additional 6.6\% relative reduction in WER over na\"{\i}ve dropout combination. 
\begin{table}[t]
  \caption{Experiments with  Stochastic dropout combination.}
  \label{stocomb}
  \centering
  \begin{tabular}{lccc}
    \toprule
    System &  \%WER (Stochastic) & \%WER (Na\"{\i}ve)\\
    \midrule
    NML-step + Forward-step &  8.76 & 8.08\\ %e152,e147, e126
    NML-step + Forward-sequence & 8.02 & 7.76\\ %e145, e124
    NML-sequence + Forward-step & 7.86 & 7.72\\ %e146, e125
    NML-sequence + Forward-sequence & 7.44 & 7.97\\ %e144, e123 (7.97-7.44)/7.97 = 0.0665
    \bottomrule
  \end{tabular}
\end{table}

Table~\ref{cascomb} illustrates one example of cascade dropout combination where we see a 4.5\% relative improvement in WER over the individual systems alone.
\begin{table}[t]
  \caption{Experiments with  Cascade dropout combination.}
  \label{cascomb}
  \centering
  \begin{tabular}{lccc}
    \toprule
    System &  \%WER  (\emph{dev-clean})\\
    \midrule
    NML-sequence + Forward-step $(1)$ &  7.72\\ %e146, e125
    NML-sequence + Forward-sequence  $(2)$& 7.97\\ %e144, e123
    Cascade  $(1) \leadsto (2)$ & 7.37\\ %e153 (7.72-7.37)/7.72 = 0.0453
    \bottomrule
  \end{tabular}
\end{table}

\section{Conclusions and future directions}

In this paper, we have described our efforts to eliminate the observed performance gap between hybrid DNN CE and LSTM-CTC ASR systems on smaller domains. We show that by combining max perturbation and stacked/strided features with feedforward and dropout  we can build an LSTM-CTC system that exceeds the performance of strong hybrid DNN CE system on a domain with 100 hours of training data. In this process, at least for the Librispeech 100hr corpus, we have 
\begin{itemize}
\item introduced max perturbation, a modified data augmentation paradigm which outperforms prior approaches.
\item shown that the combination of max perturbation, stacked/strided features, forward and recurrent dropout significantly improve LSTM-CTC ASR system performance.
\end{itemize}
Furthermore, none of the approaches in this paper explicitly make assumptions on or accommodations for corpus size, and should work in a similar manner when trained with larger datasets.

One criticism of this work is that both data augmentation and dropout would also improve the corresponding Kaldi baseline; this is most certainly the case, but even if we take the best reported results with data augmentation~\cite{KoPPK15} and dropout~\citep{ChengPPMKY17} and assume that they are additive, the LSTM-CTC system would be within 10\% relative WER of this Kaldi baseline\footnote{Ko et. al.~\cite{KoPPK15} show between 6.7\% to 0.32\% relative WER reduction using speed perturbation approach to data augmentation, Cheng et. al.~\citep{ChengPPMKY17} show 3-7\% relative WER reduction using dropout; our best system with 7.37\% WER is 6.8\% better than the Kaldi system.}, a significant leap over our initial starting point where the LSTM-CTC system was around 30-35\% worse in WER.

Another criticism of this work is that we have omitted well known approaches to improve system performance such as  discriminative training (e.g. \citep{VeselyGBP13}) and recent advances in combining different types of layers such as time delay neural network (TDNN) and convolutional layers (e.g. \cite{PeddintiPK15}) from our Kaldi baseline and thus comparing to an inferior baseline. However, evidence indicates that these approaches benefit LSTM-CTC based systems as well (e.g., \citep{AmodeiABCCCCCCD16,SakSRB15,SainathVSS15}); in \cite{VeselyGBP13} sequence discriminative training on a hybrid DNN ASR system resulted in an ~8\% relative reduction in WER whereas in \citep{SakSRB15} sequence discriminative training on a LSTM-CTC ASR system resulted in an ~10\% relative reduction in WER albeit on a different task; in \cite{PeddintiPK15}, TDNN layers are shown to result in 6.9\% relative WER reduction on the Librispeech corpus, whereas in \citep{SainathVSS15} a CLDNN (for convolution + LSTM + DNN) CE ASR system showed between 4-6\% relative WER reduction, though in later work  \citep{SeniorSQSR15}, when trained with CTC, performance was slightly worse. On the other hand, Amodei et. al. \citep{AmodeiABCCCCCCD16} use convolutional layers coupled with gated recurrent unit (GRU) layers but do not indicate the relative WER change from using only recurrent layers to the combination of convolutional and recurrent layers. 

Another issue is that we have not validated these approaches and results on other datasets to confirm that they are broadly applicable principles rather than a Librispeech corpus specific idiosyncrasy. Unfortunately, given our limited resources, we were restricted to publicly available corpora.

All said, considering the EESEN system trained on the original phoneme set as the starting point, with a 11.81\% WER, we report improvements that reduce error rates to 7.37\% WER, eliminating and improving on the performance gap with our target strong Kaldi based Hybrid DNN baseline with 7.91\% WER. While we have introduced several new approaches to improve LSTM-CTC system performance, given our limited compute pool, we have been unable to fully explore the parameter space of these improvements, i.e., we have not swept the parameters for optimal dropout rate, model size, learning rate or explored the many permutations of dropout combination.  It is likely that a more comprehensive exploration will uncover additional performance improvements.

%\subsubsection*{Acknowledgments}

%<tba>

\small

\bibliographystyle{unsrtnat}
\bibliography{LSTM-CTCperf2017}

\begin{thebibliography}{10}

\bibitem{GravesJ14}
Alex Graves and Navdeep Jaitly.
\newblock Towards end-to-end speech recognition with recurrent neural networks.
\newblock In {\em Proceedings of the 31th International Conference on Machine
  Learning, {ICML} 2014, Beijing, China, 21-26 June 2014}, pages 1764--1772,
  2014.

\bibitem{AmodeiABCCCCCCD16}
Dario Amodei, Rishita Anubhai, Eric Battenberg, Carl Case, Jared Casper, Bryan
  Catanzaro, Jingdong Chen, Mike Chrzanowski, Adam Coates, Greg Diamos, Erich
  Elsen, Jesse Engel, Linxi Fan, Christopher Fougner, Awni~Y. Hannun, Billy
  Jun, Tony Han, Patrick LeGresley, Xiangang Li, Libby Lin, Sharan Narang,
  Andrew~Y. Ng, Sherjil Ozair, Ryan Prenger, Sheng Qian, Jonathan Raiman,
  Sanjeev Satheesh, David Seetapun, Shubho Sengupta, Chong Wang, Yi~Wang,
  Zhiqian Wang, Bo~Xiao, Yan Xie, Dani Yogatama, Jun Zhan, and Zhenyao Zhu.
\newblock Deep speech 2 : End-to-end speech recognition in english and
  mandarin.
\newblock In {\em Proceedings of the 33nd International Conference on Machine
  Learning, {ICML} 2016, New York City, NY, USA, June 19-24, 2016}, pages
  173--182, 2016.

\bibitem{HannunCCCDEPSSCN14}
Awni~Y. Hannun, Carl Case, Jared Casper, Bryan Catanzaro, Greg Diamos, Erich
  Elsen, Ryan Prenger, Sanjeev Satheesh, Shubho Sengupta, Adam Coates, and
  Andrew~Y. Ng.
\newblock Deep speech: Scaling up end-to-end speech recognition.
\newblock {\em CoRR}, abs/1412.5567, 2014.

\bibitem{SakSRB15}
Hasim Sak, Andrew~W. Senior, Kanishka Rao, and Fran{\c{c}}oise Beaufays.
\newblock Fast and accurate recurrent neural network acoustic models for speech
  recognition.
\newblock In {\em {INTERSPEECH} 2015, 16th Annual Conference of the
  International Speech Communication Association, Dresden, Germany, September
  6-10, 2015}, pages 1468--1472, 2015.

\bibitem{SoltauLS16}
Hagen Soltau, Hank Liao, and Hasim Sak.
\newblock Neural speech recognizer: Acoustic-to-word {LSTM} model for large
  vocabulary speech recognition.
\newblock {\em CoRR}, abs/1610.09975, 2016.

\bibitem{GravesFGS06}
Alex Graves, Santiago Fern{\'{a}}ndez, Faustino~J. Gomez, and J{\"{u}}rgen
  Schmidhuber.
\newblock Connectionist temporal classification: labelling unsegmented sequence
  data with recurrent neural networks.
\newblock In {\em Machine Learning, Proceedings of the Twenty-Third
  International Conference {(ICML} 2006), Pittsburgh, Pennsylvania, USA, June
  25-29, 2006}, pages 369--376, 2006.

\bibitem{PoveyPGGMNWK16}
Daniel Povey, Vijayaditya Peddinti, Daniel Galvez, Pegah Ghahremani, Vimal
  Manohar, Xingyu Na, Yiming Wang, and Sanjeev Khudanpur.
\newblock Purely sequence-trained neural networks for {ASR} based on
  lattice-free {MMI}.
\newblock In {\em Interspeech 2016, 17th Annual Conference of the International
  Speech Communication Association, San Francisco, CA, USA, September 8-12,
  2016}, pages 2751--2755, 2016.

\bibitem{MiaoGM15}
Yajie Miao, Mohammad Gowayyed, and Florian Metze.
\newblock {EESEN:} end-to-end speech recognition using deep {RNN} models and
  wfst-based decoding.
\newblock In {\em 2015 {IEEE} Workshop on Automatic Speech Recognition and
  Understanding, {ASRU} 2015, Scottsdale, AZ, USA, December 13-17, 2015}, pages
  167--174, 2015.

\bibitem{PoveyASRU2011}
Daniel Povey, Arnab Ghoshal, Gilles Boulianne, Lukas Burget, Ondrej Glembek,
  Nagendra Goel, Mirko Hannemann, Petr Motlicek, Yanmin Qian, Petr Schwarz, Jan
  Silovsky, Georg Stemmer, and Karel Vesely.
\newblock The kaldi speech recognition toolkit.
\newblock In {\em IEEE 2011 Workshop on Automatic Speech Recognition and
  Understanding}. IEEE Signal Processing Society, December 2011.
\newblock IEEE Catalog No.: CFP11SRW-USB.

\bibitem{PanayotovCPK15}
Vassil Panayotov, Guoguo Chen, Daniel Povey, and Sanjeev Khudanpur.
\newblock Librispeech: An {ASR} corpus based on public domain audio books.
\newblock In {\em 2015 {IEEE} International Conference on Acoustics, Speech and
  Signal Processing, {ICASSP} 2015, South Brisbane, Queensland, Australia,
  April 19-24, 2015}, pages 5206--5210, 2015.

\bibitem{SchusterP97}
Mike Schuster and Kuldip~K. Paliwal.
\newblock Bidirectional recurrent neural networks.
\newblock {\em {IEEE} Trans. Signal Processing}, 45(11):2673--2681, 1997.

\bibitem{GersSC00}
Felix~A. Gers, J{\"{u}}rgen Schmidhuber, and Fred~A. Cummins.
\newblock Learning to forget: Continual prediction with {LSTM}.
\newblock {\em Neural Computation}, 12(10):2451--2471, 2000.

\bibitem{JozefowiczZS15}
Rafal J{\'{o}}zefowicz, Wojciech Zaremba, and Ilya Sutskever.
\newblock An empirical exploration of recurrent network architectures.
\newblock In {\em Proceedings of the 32nd International Conference on Machine
  Learning, {ICML} 2015, Lille, France, 6-11 July 2015}, pages 2342--2350,
  2015.

\bibitem{MiaoGNKMW16}
Yajie Miao, Mohammad Gowayyed, Xingyu Na, Tom Ko, Florian Metze, and
  Alexander~H. Waibel.
\newblock An empirical exploration of {CTC} acoustic models.
\newblock In {\em 2016 {IEEE} International Conference on Acoustics, Speech and
  Signal Processing, {ICASSP} 2016, Shanghai, China, March 20-25, 2016}, pages
  2623--2627, 2016.

\bibitem{KoPPK15}
Tom Ko, Vijayaditya Peddinti, Daniel Povey, and Sanjeev Khudanpur.
\newblock Audio augmentation for speech recognition.
\newblock In {\em {INTERSPEECH} 2015, 16th Annual Conference of the
  International Speech Communication Association, Dresden, Germany, September
  6-10, 2015}, pages 3586--3589, 2015.

\bibitem{jaitly2013}
Navdeep Jaitly and G.~E. Hinton.
\newblock Vocal tract length perturbation (vtlp) improves speech recognition.
\newblock In {\em Proceedings of the International Conference on Machine
  Learning (ICML) 2013 Workshop on Deep Learning for Audio, Speech and Language
  Processing, Atlanta, Georgia, USA, June 16-21, 2013}, 2013.

\bibitem{CuiGK15}
Xiaodong Cui, Vaibhava Goel, and Brian Kingsbury.
\newblock Data augmentation for deep convolutional neural network acoustic
  modeling.
\newblock In {\em 2015 {IEEE} International Conference on Acoustics, Speech and
  Signal Processing, {ICASSP} 2015, South Brisbane, Queensland, Australia,
  April 19-24, 2015}, pages 4545--4549, 2015.

\bibitem{Kirkpatrick1983}
S.~Kirkpatrick, C.~D. Gelatt, and M.~P. Vecchi.
\newblock Optimization by simulated annealing.
\newblock {\em Science}, 220(4598):671--680, 1983.

\bibitem{SakSB14}
Hasim Sak, Andrew~W. Senior, and Fran{\c{c}}oise Beaufays.
\newblock Long short-term memory recurrent neural network architectures for
  large scale acoustic modeling.
\newblock In {\em {INTERSPEECH} 2014, 15th Annual Conference of the
  International Speech Communication Association, Singapore, September 14-18,
  2014}, pages 338--342, 2014.

\bibitem{SrivastavaHKSS14}
Nitish Srivastava, Geoffrey~E. Hinton, Alex Krizhevsky, Ilya Sutskever, and
  Ruslan Salakhutdinov.
\newblock Dropout: a simple way to prevent neural networks from overfitting.
\newblock {\em Journal of Machine Learning Research}, 15(1):1929--1958, 2014.

\bibitem{ZarembaSV14}
Wojciech Zaremba, Ilya Sutskever, and Oriol Vinyals.
\newblock Recurrent neural network regularization.
\newblock {\em CoRR}, abs/1409.2329, 2014.

\bibitem{MoonCLS15}
Taesup Moon, Heeyoul Choi, Hoshik Lee, and Inchul Song.
\newblock {RNNDROP:} {A} novel dropout for {RNNS} in {ASR}.
\newblock In {\em 2015 {IEEE} Workshop on Automatic Speech Recognition and
  Understanding, {ASRU} 2015, Scottsdale, AZ, USA, December 13-17, 2015}, pages
  65--70, 2015.

\bibitem{SemeniutaSB16}
Stanislau Semeniuta, Aliaksei Severyn, and Erhardt Barth.
\newblock Recurrent dropout without memory loss.
\newblock In {\em {COLING} 2016, 26th International Conference on Computational
  Linguistics, Proceedings of the Conference: Technical Papers, December 11-16,
  2016, Osaka, Japan}, pages 1757--1766, 2016.

\bibitem{GalG16}
Yarin Gal and Zoubin Ghahramani.
\newblock A theoretically grounded application of dropout in recurrent neural
  networks.
\newblock In {\em Advances in Neural Information Processing Systems 29: Annual
  Conference on Neural Information Processing Systems 2016, December 5-10,
  2016, Barcelona, Spain}, pages 1019--1027, 2016.

\bibitem{PhamBKL14}
Vu~Pham, Th{\'{e}}odore Bluche, Christopher Kermorvant, and J{\'{e}}r{\^{o}}me
  Louradour.
\newblock Dropout improves recurrent neural networks for handwriting
  recognition.
\newblock In {\em 14th International Conference on Frontiers in Handwriting
  Recognition, {ICFHR} 2014, Crete, Greece, September 1-4, 2014}, pages
  285--290, 2014.

\bibitem{ChengPPMKY17}
Gaofeng Cheng, Vijayaditya Peddinti, Daniel Povey, Vimal Manohar, Sanjeev
  Khudanpur, and Yonghong Yan.
\newblock An exploration of dropout with lstms.
\newblock In {\em {INTERSPEECH} 2017, Annual Conference of the International
  Speech Communication Association, Stockholm, Sweden, August 20-24, 2017},
  pages xxxx--xxxx, 2017.

\bibitem{VeselyGBP13}
Karel Vesel{\'{y}}, Arnab Ghoshal, Luk{\'{a}}s Burget, and Daniel Povey.
\newblock Sequence-discriminative training of deep neural networks.
\newblock In Fr{\'{e}}d{\'{e}}ric Bimbot, Christophe Cerisara, C{\'{e}}cile
  Fougeron, Guillaume Gravier, Lori Lamel, Fran{\c{c}}ois Pellegrino, and
  Pascal Perrier, editors, {\em {INTERSPEECH} 2013, 14th Annual Conference of
  the International Speech Communication Association, Lyon, France, August
  25-29, 2013}, pages 2345--2349. {ISCA}, 2013.

\bibitem{PeddintiPK15}
Vijayaditya Peddinti, Daniel Povey, and Sanjeev Khudanpur.
\newblock A time delay neural network architecture for efficient modeling of
  long temporal contexts.
\newblock In {\em {INTERSPEECH} 2015, 16th Annual Conference of the
  International Speech Communication Association, Dresden, Germany, September
  6-10, 2015}, pages 3214--3218. {ISCA}, 2015.

\bibitem{SainathVSS15}
Tara~N. Sainath, Oriol Vinyals, Andrew~W. Senior, and Hasim Sak.
\newblock Convolutional, long short-term memory, fully connected deep neural
  networks.
\newblock In {\em 2015 {IEEE} International Conference on Acoustics, Speech and
  Signal Processing, {ICASSP} 2015, South Brisbane, Queensland, Australia,
  April 19-24, 2015}, pages 4580--4584. {IEEE}, 2015.

\bibitem{SeniorSQSR15}
Andrew~W. Senior, Hasim Sak, Felix de~Chaumont~Quitry, Tara~N. Sainath, and
  Kanishka Rao.
\newblock Acoustic modelling with {CD-CTC-SMBR} {LSTM} {RNNS}.
\newblock In {\em 2015 {IEEE} Workshop on Automatic Speech Recognition and
  Understanding, {ASRU} 2015, Scottsdale, AZ, USA, December 13-17, 2015}, pages
  604--609, 2015.

\end{thebibliography}

\end{document}